\documentclass[10pt,a4paper]{article}

\usepackage{authblk}
\usepackage[margin=2.5cm]{geometry}
\usepackage[utf8]{inputenc} 
\usepackage[T1]{fontenc}    
\usepackage{hyperref}       
\usepackage{url}            
\usepackage{booktabs}       
\usepackage{amsfonts}       
\usepackage{nicefrac}       
\usepackage{microtype}      
\usepackage{xcolor}         

\usepackage{graphicx} 
\usepackage{amssymb}
\usepackage{amsmath}
\usepackage{mathtools}  
\usepackage{algorithmic}
\usepackage{subcaption}
\usepackage{amsthm}
\usepackage{ifthen}
\usepackage{adjustbox}

\newcommand \be {\begin{equation}}
\newcommand \ee {\end{equation}}
\newcommand \RR    {\mathbb{R}}

\newcommand \NN    {\mathbb{N}}

\newcommand{\inner}[2]{\ifthenelse{\equal{#2}{}}{\left\langle\cdot,\cdot\right\rangle_{#1}}{\left\langle#2\right\rangle_{#1}}}
\newcommand{\norm}[2]{\ifthenelse{\equal{#2}{}}{\left\|\cdot\right\|_{#1}}{\left\|#2\right\|_{#1}}}

\usepackage{enumitem}

\title{Differentiable Kernel Ridge Regression for Deep Learning Pipelines}

\author[1]{Jean-Marc Mercier\thanks{jeanmarc.mercier@gmail.com}}
\author[2]{Gabriele Santin\thanks{gabriele.santin@unive.it}}

\affil[1]{MPG-Partners, Paris, France}
\affil[2]{Department of Environmental Sciences, Informatics and Statistics, Ca' Foscari University of Venice, Italy}

\date{\today}

\begin{document}

\maketitle

\begin{abstract}
Deep neural networks dominate modern machine learning, while alternative function approximators remain comparatively underexplored at scale. In this work, we 
revisit kernel methods as drop-in components for standard deep learning pipelines. We introduce \emph{Sparse Kernels} (SKs), a differentiable, localized, and 
lazy variant of kernel ridge regression (KRR) that defers training to inference time and reduces to the solution of small local systems. We integrate SKs into 
PyTorch as modular layers that preserve end-to-end trainability, and we show that they expose three distinct sets of parameters---feature representations, 
target values, and evaluation points---each of which can be fixed or learned. This decomposition broadens the design space available to practitioners, enabling, 
in particular, training-free transfer, nonlinear probing, and hybrid kernel-neural models. Across convolutional networks, vision transformers, and reinforcement 
learning, SK-based modules serve two complementary roles: in some settings, they match the performance of trained neural readouts with substantially less 
training; in others, they augment existing models and improve their performance when used as additional components. Our results suggest that kernel methods, 
once made scalable and differentiable, can be readily integrated with deep learning rather than treated as a separate paradigm.
\end{abstract}

\section{Introduction}

The goal of this paper is to study the interplay between deep kernel learning~\cite{Salakhutdinov:2016} and localized kernel methods~\cite{Han:2022} for 
scalable machine learning, and to make the resulting framework directly usable inside modern deep learning pipelines.

Deep kernel learning combines neural networks with kernel methods by learning data-dependent feature representations and applying kernel-based predictors---such 
as kernel ridge regression (KRR) or Gaussian processes---in the resulting feature space~\cite{Wilson:2015}. This approach leverages the expressive power of deep 
architectures while retaining the theoretical grounding and flexibility of kernel methods, including nonparametric modeling and uncertainty quantification. A 
key limitation of classical kernel methods, however, is their computational cost, which restricts their applicability at scale.

Recent work on local kernel ridge regression mitigates this issue by restricting predictions to localized neighborhoods~\cite{Han:2022,Schaback:2025}. Building 
on this idea, we introduce \emph{sparse kernels} (SKs), a lazy variant of local KRR\footnote{We use ``lazy'' in the sense of deferred computation, not in the 
sense of the lazy/NTK training regime studied in the infinite-width literature.} in which training is effectively deferred to inference time, requiring only the 
solution of small local systems. This yields a scalable learning paradigm that preserves desirable properties such as consistency and continuity while 
substantially reducing computational complexity. As a result, SKs provide a practical instantiation of localized KRR that can be seamlessly integrated into 
modern deep learning pipelines.

A central perspective of our approach is that KRR naturally decomposes learning into three interacting components: feature representations, target values, and 
evaluation points. Each of these components can be treated as fixed or learnable, enabling alternatives to standard end-to-end training, including training-free 
inference for transfer learning and nonlinear probing of intermediate representations.

This perspective also lets us revisit the role of the readout layer in deep models. While linear probes are commonly used to assess learned 
representations~\cite{Bengio:2016}, they are restricted to linear function classes. By contrast, SK-based probes provide nonlinear, data-adaptive readouts that 
better capture the geometry of learned features. More broadly, our approach echoes recent work showing that kernel methods can be competitive with neural 
networks when applied to fixed representations~\cite{Ghorbani:2020, Lee:2020}. Our results further show that this observation extends beyond probing to fully 
trained models.

We provide initial empirical evidence of this approach through experiments on convolutional networks, vision transformers, and reinforcement learning tasks. 
SK-based models achieve competitive performance while reducing training requirements, and kernel-based readouts exploit learned representations more effectively 
than standard parametric heads, particularly in transformer architectures.

\subsection*{Related work}\label{RelatedWork}
\paragraph{Deep kernel learning and infinite-width limits.} Combining neural feature extractors with kernel-based predictors has a long history. Deep Kernel 
Learning~\cite{Salakhutdinov:2016, Wilson:2015} learns data-dependent representations and applies a kernel method (typically a Gaussian process) on top. In a 
different direction, the Neural Tangent Kernel viewpoint and related infinite-width analyses~\cite{Ghorbani:2020,Lee:2020} characterize when finite networks 
behave like kernel machines. Our work is complementary: rather than studying the limit equivalence, we provide a practical, differentiable kernel module that 
can be slotted into finite, trainable architectures, showing that the hybrid models remain competitive without retraining.

\paragraph{Localized and structured kernel methods.} Scalability of kernel methods has been pursued through inducing-point approximations such as 
KISS-GP~\cite{Wilson:2015}, structured deep kernels~\cite{Santin:2025}, and local kernel ridge regression~\cite{Han:2022}. Our sparse kernels build directly on 
the latter, using $M$-NN to define a tessellation and deferring kernel solves to inference time. Compared with~\cite{Han:2022}, we propose a discontinuous 
kernel extrapolation for performance reasons, a differentiable implementation that integrates with deep learning backends, and we expose evaluation points, 
features, and targets as independently controllable parameter sets.

\paragraph{Probing.} Linear probes~\cite{Bengio:2016} are a standard tool for analyzing learned representations. Recent work generalizes probing to more 
expressive predictors~\cite{Adams:2025} or uses transfer with ridge regression~\cite{Tang:2020}. Our kernel probes provide a nonlinear, data-adaptive readout 
that, unlike a learned MLP head, requires no additional training in the lazy regime and exposes geometric notions of representation quality from the RKHS 
structure, such as the local error indicator $\epsilon_k^\sigma$ (see Appendix~\ref{SparseKernelErrorEstimates}).

\subsection*{Contributions}\label{Contributions} Our work introduces three main contributions:
\begin{itemize}
    \item \textbf{Localized lazy KRR.} We formalize a local, differentiable, lazy and sparse variant of kernel ridge regression in the interpolating regime, 
clarify its relationship to global KRR and nearest-neighbor methods, and adapt classical KRR error estimates to this localized setting 
(Appendix~\ref{SparseKernelErrorEstimates}).
    
    \item \textbf{A modular kernel framework for deep learning.} We integrate a kernel library~\cite{LeMeMi:2025} as dense or sparse kernel modules within a 
standard deep learning framework, exposing three sets of parameters---representations, targets, and evaluation points---that can be fixed or learned and enable 
new training strategies.
    
    \item \textbf{Empirical validation.} We demonstrate the effectiveness of the proposed approach across multiple domains, including transfer learning with 
training-free inference, nonlinear probing of pretrained models, and a kernel-augmented variant of Double DQN in reinforcement learning.
\end{itemize}

\paragraph{Notation.}

We let $x=(x^1,\dots,x^{N})\in\RR^{N \times D_x}$ be a matrix, with rows $x^n=(x^n_1,\dots,x^n_{D_x})\in\RR^{D_x}$.
The $\ell^p$ norm is $|x^n|_p = \big(\sum_{d=1}^{D_x} |x^n_d|^p\big)^{1/p}$ for vectors and $\|x\|_p = \big(\sum_{n=1}^{N} |x^n|_p^p\big)^{1/p}$ for matrices.
For $\bar x\in\RR^{N \times D_x}$, we write $\langle x,\bar x\rangle = \sum_{n,d} x^n_d \bar x^n_d$ for the standard $\ell_2$ inner product.
We use vectorized notation for function evaluations: if $y:\RR^{D_x}\to\RR^{D_y}$ then $y(x) =(y(x^1),\dots,y(x^{N})) \in \RR^{N \times D_y}$, while for 
$k:\RR^{D_x}\times \RR^{D_x}\to\RR$ we have $k(\cdot,x):\RR^{D_x}\to \RR^{N}$ and $k(x,x)=(k(x^i,x^j))_{i,j} \in \RR^{N \times N}$.
If $M\leq N$ and $\sigma=(\sigma^1,\dots,\sigma^M) \in \{1,\dots, N\}^M$ is a tuple of indices, then $x^\sigma=(x^{\sigma^1},\dots,x^{\sigma^M})$ denotes the 
corresponding subset.
We let $\Sigma_{M}^{N} = \{\sigma = (\sigma^1, \dots, \sigma^M) \in \{1,\dots,N\}^M : \sigma^n\neq\sigma^m \text{ if } n \neq m\}$ be the set of $M$-tuples of 
distinct indices, abbreviated as $\Sigma$ when no ambiguity arises.
Finally, $I_N \in \RR^{N \times N}$ is the identity matrix, $\mathbf{1}_\Omega(x)$ is the indicator function of a set $\Omega$, and $\mathrm{ReLU}(a)=\max(a,0)$ 
denotes the rectified linear activation.

\section{Deep kernel methods}\label{Reminder-on-RKHS-methods}
\subsection{Dense kernel ridge regression}\label{Dense-Kernel-Ridge-Regression}
We recall the basics of kernel ridge regression~\cite{Wendland:2005,Scholkopf:2002} that we need to describe deep kernel approaches and sparse kernels. Given a 
finite set of $N$ distinct features $x=(x^1,\dots,x^{N}) \in \RR^{N \times D_x}$ with target values $y=(y^1,\dots,y^{N}) \in \RR^{N \times D_y}$, we consider a 
strictly positive definite kernel, that is, a symmetric function $k:\RR^{D_x} \times \RR^{D_x} \to \RR$ such that the Gram matrix $k(x,x) \in \RR^{N \times N}$ 
is positive definite.
This kernel defines a unique Reproducing Kernel Hilbert Space (RKHS) $\mathcal{H}_{k}$, and also a finite-dimensional space of vector-valued functions 
parameterized by points $x$ and matrices $\theta \in \RR^{N \times D_y}$:
\be
\label{Hk}
\mathcal{H}_{k}^{x} = \Bigl\{
y_{k}(\cdot,\theta) = \sum_{n=1}^{N} \theta^n k(\cdot,x^n) = k(\cdot,x) \theta, \quad \theta \in \RR^{N \times D_y} \Bigr\}.
\ee
Letting $y_{k,\theta} \coloneqq y_{k}(x,\theta)=k(x,x)\theta\in\RR^{N \times D_y}$, the space $\mathcal{H}_{k}^{x}$ is equipped with the inner product induced 
by $\|y_{k}(\cdot,\theta)\|^2_{\mathcal{H}_{k}^{x}} = \langle \theta, y_{k,\theta} \rangle = \theta^T k(x,x)\theta$.

We work in the \emph{interpolation regime} by assuming that the kernel is strictly positive definite, so that $\dim(\mathcal{H}_{k}^{x})=N$ and the matrix 
$k(x,x)$ is invertible without any further regularization (e.g.\ Tikhonov).
This induces a set of $N$ \emph{cardinal} (or \emph{Lagrange}) basis functions of $\mathcal{H}_{k}^{x}$, defined by
\be
\label{psi}
\psi_k(\cdot)= (\psi_k^1(\cdot),\dots,\psi_k^{N}(\cdot)) = k(\cdot,x)\,k(x,x)^{-1} \in \RR^{N},
\ee
which satisfy $\psi_k^n(x^m) = \delta_{n,m}$ and parameterize $\mathcal{H}_{k}^{x}$ via function values at $x$ as
\be
\label{Hky}
\mathcal{H}_{k}^{x} = \Bigl\{
y_k(\cdot, y) = \sum_{n=1}^{N} y^n \psi_k^n(\cdot) = \psi_k(\cdot)\, y, \quad y \in \RR^{N \times D_y} \Bigr\},\;\;\|y_k(\cdot, y)\|^2_{\mathcal{H}_{k}^{x}} = 
y^T k(x,x)^{-1} y.
\ee

To summarize, given a dataset $x,y$, this construction defines an extrapolation operator $z \mapsto \mathcal{P}_k(x,y)(z)$, called the \emph{kernel ridge 
regressor}, which we denote by $y_k(z)$ or $y_k(z,x,y)$ when the dependence on its parameters needs to be made explicit. The map is continuous everywhere if $k$ 
is, and in the interpolation regime it satisfies $y_k(x,x,y)=y$.
Kernel methods come with several local, global, and confidence-interval-type error bounds~\cite{singh2023kernel}. 

\subsection{Deep learning approaches}
\paragraph{Deep learning.}
To learn an association $x \mapsto y$, modern models are typically composed of $L$ layers, starting from $y^0 = x \in \RR^{N \times D_x}$ and producing the 
target $y^{L} = y \in \RR^{N \times D_y}$.
To each layer $l=1,\dots,L$ we attach a parameter set $\theta^l$ and a function $y^l(\cdot,\theta^l)$. Letting $\theta = (\theta^1,\dots,\theta^{L})$ denote the 
full parameter set, we have
\be \label{deep}
 y^{l}_{\theta}=y^{l}(y^{l-1}_{\theta},\theta^{l}) \in \RR^{N \times D^l}, \qquad l=1,\dots,L,
\ee
for some $D^l$. A loss functional $\mathcal{L}(\theta)$ is associated with this construction, and parameters are obtained by approximately solving $\inf_\theta 
\mathcal{L}(\theta)$, typically via stochastic gradient or quasi-Newton optimizers combined with automatic differentiation. In this context, we propose a 
kernel-based implementation of a single layer that integrates with a widely adopted backend, namely PyTorch~\cite{Paszke:2019}, which we describe below as a 
two-layer analysis.

\paragraph{Two-layer analysis.}
We focus on a two-layer model that learns an association $x \mapsto y$, mapping $\RR^{D_x} \to \RR^{D_y}$ via $x \mapsto f(x,\theta) \mapsto y$, with the 
intermediate space $\RR^{D_f}$. This decomposition splits the model into \emph{feature} and \emph{classifier} mappings.

\textit{Feature map.} A feature map is a mapping $f(\cdot,\theta) :  \RR^{D_x} \to \RR^{D_{f}}$ with parameters $\theta$. In the present analysis, both 
$f(\cdot,\theta)$ and its derivatives $\nabla_{\theta} f(\cdot,\theta)$ are assumed to be available, e.g.\ through a third-party automatic differentiation 
engine.

\textit{Classifier map.} A classifier map is a mapping  $y(\cdot,\theta_y) : \RR^{D_{f}} \to \RR^{D_y}$ with parameters $\theta_y$. The full forward two-layer 
model is then
\be  \label{NN}
y_{\theta,\theta_y}(\cdot)=y(f(\cdot,\theta),\theta_y).
\ee

For example, in a convolutional network such as VGG~\cite{Simonyan:2014}, $\theta$ contains the convolutional filter parameters, $f(\cdot,\theta)$ is the 
composition of all layers up to the classifier, and the classifier itself has its own parameters $\theta_y$ in~\eqref{NN}.

\paragraph{Kernel classifier map.} Replacing the classifier in~\eqref{NN} by a kernel ridge regressor introduces two additional parameter sets 
$\theta_x,\theta_y$, yielding the fully parameterized two-layer kernel model
\be \label{2L}
    y_{\theta}(\cdot, \theta_x,\theta_y)=y_{k}(f(\cdot,\theta),\theta_x, \theta_y),
\ee
where $y_k(\cdot,\theta_x,\theta_y)= k(\cdot,\theta_x)\,k(\theta_x,\theta_x)^{-1}\theta_y$ is the kernel ridge regressor. Several choices are available for 
these kernel parameters. First, equation~\eqref{2L} can be used without introducing extra parameters by using propagated points $\theta_x=x_{\theta} \coloneqq 
f(x,\theta) \in \RR^{N \times D_f}$ and $\theta_y=y \in \RR^{N \times D_y}$,
\be \label{2LNOPARAM}
    y_{\theta}(\cdot)= k(f(\cdot,x_\theta),x_\theta)\,k(x_\theta,x_\theta)^{-1}\,y,
\ee
which is appropriate when $y$ is observed, as typical for a classification model.
On the other hand, in several scenarios of interest, where the targets are learnable, we let $y_{\theta}(\cdot,\theta_y)= 
k(f(\cdot,x_\theta),x_\theta)\,k(x_\theta,x_\theta)^{-1}\theta_y$.
This paper explores in Section~\ref{ReinforcementLearning} a hybrid configuration in which both $\theta_x$ and $\theta_y$ are treated as learnable parameters, 
as in the fully parameterized model~\eqref{2L}. Note that loss terms tailored to kernel methods can be considered to learn these parameters: a natural candidate 
is an optimal-transport-based regularizer (see Appendix~\ref{OT}).

\paragraph{Algorithmic complexity.} The cost of evaluating the kernel classifier~\eqref{2L} is
\be \label{CA}
    \mathcal{O}(N^3) + \mathcal{O}(N^2\, D_y) \text{ operations},
\ee
where the cubic term comes from the matrix inversion and the quadratic term from matrix multiplication. With minibatch optimizers, $N$ is the batch size; for 
typical small batch sizes, the dense formulation of Section~\ref{Dense-Kernel-Ridge-Regression} suffices, while larger batches benefit from the sparse 
formulation of Section~\ref{Sparse-kernel-methods}. In a fair benchmark, the cost~\eqref{CA} should be compared not only against the cost of evaluating the 
competing classifier in~\eqref{NN}, but also against the cost of the feature map $f(\cdot,\theta)$. Empirically, training a model with a neural classifier or a 
kernel classifier is comparable for standard batch sizes.

We now describe the loss-based functional setting commonly used in deep learning frameworks.

\subsubsection{Representer theorem and functionals}

For $a,b \in \RR^{M \times D}$ we let $\ell:\RR^{D} \times \RR^{D}\to\RR$ be a per-sample loss (e.g.\ cross-entropy or mean squared error). We consider in this 
paper aggregate losses of the form
$\mathcal{L}(a,b) \coloneqq \sum_{n=1}^{M} \ell(a^n,b^n)$.
Let $z \in \RR^{N_z \times D_z}$ be evaluation features mapped to target values $y_z \in \RR^{N_z \times D_y}$. For a two-layer model with a kernel classifier 
and the shorthand $x_\theta=f(x,\theta)$, $z_\theta=f(z,\theta)$, $y_{\theta}=y_k\big(z_\theta,x_\theta\big)$ (see~\eqref{2L}), the loss reads
\be \label{fun}
\mathcal{L}(y_{\theta},y_z)
\coloneqq
\sum_{n=1}^{N_z} \ell \big(y_{\theta}^n,y_z^n\big),
\ee
and the training problem is
$
\inf_{y,\theta}\; \mathcal{L}(\mathcal{P}_k(x_\theta,y),y_z)
$.
From a theoretical standpoint, the representer theorem \cite{Santin:2025,Scholkopf:2002} ensures the equivalence between the discrete problem $\inf_{y,\theta} 
\mathcal{L}(y_{\theta},y_z)$ and the functional minimization $\inf_{y_k(\cdot) \in \mathcal{H}_{k}^{x_\theta}} \mathcal{L}(y_k(z,x_\theta),y_z)$.

\paragraph{Implementation.} Most deep learning frameworks have their own optimizers, such as Adam, SGD or LBFGS. They rely on gradient computations during the 
descent step, which in our setting amount to
$
\nabla_y\, y_k(z,x,y)$, $\nabla_x\, y_k(z,x,y)$, $\nabla_z\, y_k(z,x,y)
$,
that is, the derivatives of the kernel ridge regressor with respect to each of its three arguments. These derivatives are computed by the kernel library 
of~\cite{LeMeMi:2025} and exported to the deep learning backend.

\section{Sparse kernel methods}\label{Sparse-kernel-methods}
\subsection{$M$-NN (Nearest Neighbors)}

Throughout this section we rely on an $M$-nearest-neighbor approach, since efficient GPU-distributed implementations are publicly available.
We then build our sparse kernel construction directly on top of this $M$-NN structure, without subsampling the dataset. When subsampling is required, $M$-NN 
also provides an efficient way to select a representative subset for KRR, as detailed in Appendix~\ref{GreedySelection}.

\subsubsection{$M$-NN tessellation}
Let $d(\cdot,\cdot)$ be a distance, and $x \in \mathbb{R}^{N,D_x}$ a set of points. Let $1 \le M \le N$ be a \textit{bandwidth} parameter, and consider the 
local ordering $\sigma(\cdot)=(\sigma^1,\cdots,\sigma^M)(\cdot) \in \Sigma$ given by the combination satisfying on any given point $z \in \mathbb{R}^{D_x}$
$$
  d(z, x^{\sigma^n(z)}) \le d(z, x^{\sigma^m(z)}), \quad n \le m \le M.
$$
Consider the function $z \mapsto \sigma(z) =(\sigma^1,\cdots,\sigma^M)(z)$, the first $M$-truncation of the nearest-neighbor ordering at a point $z$. This 
function is piecewise constant and defines a partition (or tessellation) of the space $\RR^{D_x}$, denoted $\mathbf{1}_{\Omega^\sigma}$, into small regions 
sharing the same $M$-neighborhood:
$$
  1 = \sum_{\sigma \in \Sigma} \mathbf{1}_{\Omega^\sigma}(x), \quad \Omega^{\sigma} := \{x \in \RR^{D_x} : \sigma(x) = \sigma\},
$$
to which we associate $x^{\sigma(z)} = (x^{\sigma^1(z)}, \cdots,x^{\sigma^M(z)})$, the $M$-truncated vector of features. In what follows, we write $\sigma = 
\sigma(\cdot) = (\sigma^n(\cdot))_{n=1,\cdots,M}$ for concision. For example, the evaluation of the indices on the training set $x$ is $\sigma(x) \in 
\NN^{N,M}$.

\subsection{Sparse, Local, Reproducing Kernel Hilbert Spaces}
With $x=(x^1,\dots,x^{N})$ as input features and $y=(y^1,\dots,y^{N})$ as targets, we fix a positive definite kernel of the form $k(\cdot,\cdot) = 
\varphi(d(\cdot,\cdot))$ for some distance $d$ and a positive definite function (activation) $\varphi$~\cite{Wendland:2005}, and let $\sigma(\cdot)$ denote the 
$M$-NN truncation introduced above. On each cell $\Omega^\sigma$ we define
\be
\label{RHk}
\mathcal{H}_{k,\sigma} = \Bigl\{
y_k(\cdot) = \psi_k(\cdot,x^{\sigma})\, y^{\sigma},\quad y^\sigma \in \RR^{M \times D_y} \Bigr\},
\ee
where $\sigma = \sigma(\cdot)$ is the local cell to which the evaluation point belongs and
$$\psi_k(\cdot,x^\sigma) = k(\cdot,x^\sigma)\,k(x^\sigma,x^\sigma)^{-1}$$
are the local cardinal basis functions, with $k(x^\sigma,x^\sigma)$ a \textbf{dense} matrix of size $M \times M$. The number of free parameters per cell is $M 
\times D_y$. We also note that kernels typically come with normalizing maps that may themselves depend on the cell $\Omega^\sigma$.

The indicator family $\mathbf{1}_{\Omega^\sigma}$ is a partition of unity, so the direct sum $\bigoplus_{\sigma \in \Sigma} \mathcal{H}_{k,\sigma}$ can be 
equipped with the norm $\sum_{\sigma}\|y\, \mathbf{1}_{\Omega^\sigma}\|_{\mathcal{H}_{k,\sigma}}^2$. This is a finite-dimensional space of functions that are 
continuous on each cell $\Omega^\sigma$ but possibly discontinuous globally; it has $\theta \in \RR^{\#\Sigma \times M \times D_y}$ free parameters, where 
$\#\Sigma$ is the number of cells induced by $x$.

We additionally consider the functional space
\be \label{SK}
\mathcal{H}_{k}^{M} = \Bigl\{
y_k(z) = \psi_k(z,x^\sigma)\, y^\sigma,\quad y \in \RR^{N \times D_y},\ z \in \Omega^{\sigma} \Bigr\},
\ee
where the parameters $y \in \RR^{N \times D_y}$ are now defined globally  but act only locally on each cell, through $y^{\sigma}$. $\mathcal{H}_{k}^M$ is a 
closed subspace of $\bigoplus_{\sigma \in \Sigma} \mathcal{H}_{k,\sigma}$, hence inherits the norm $\|y\|_{\mathcal{H}_{k}^M}^2 = \sum_{\sigma \in \Sigma}\|y\, 
\mathbf{1}_{\Omega^\sigma}\|_{\mathcal{H}_{k,\sigma}}^2$.

When $k$ is positive definite, this construction yields a kernel ridge regressor whose cost is linear in the dataset size $N$ and which has the following 
properties:
\begin{itemize}
\item It defines a piecewise continuous function with respect to the partition $\mathbf{1}_{\Omega^\sigma}$, although it may be globally discontinuous. Our 
experiments use this discontinuous formulation, which performs well in practice. Globally continuous variants can also be defined; see 
Appendix~\ref{HierarchicalContinuousSparseKernelRegressors}, which is close in spirit to the construction of~\cite{Han:2022}.
\item Classical error estimates available for RKHS kernel ridge regression can be easily adapted to this SK setting, see Appendix section 
\ref{SparseKernelErrorEstimates}. 
\item Since the basis functions $\psi_k(z,x^\sigma)$ are computed with respect to the evaluation point $z$, the method is effectively \emph{lazy}: most of its 
cost is paid at inference time. Each prediction is independent and requires $\mathcal{O}(M^3)$ operations, with accuracy controlled by the local error bound 
~\eqref{EBT} in the appendix A.2.
\end{itemize}

\section{Numerical experiments}

For a fair comparison, all benchmarks use the same features and the same data splits. Experimental details are reported in the Appendix, Section 
\ref{NumericalExperimentSettings}.

\subsection{Transfer learning}\label{Transfer}
We evaluate four transfer learning strategies on top of a ResNet-18~\cite{He:2016} pretrained on ImageNet~\cite{Deng:2009}. The feature extractor is kept 
frozen, and we compare four readout mechanisms over varying subsample sizes of the CIFAR-10 validation set: (i-Linear Probe in figure) a linear classification 
head, (ii - MLP(512-10)) a multilayer perceptron (MLP), (iii - SK(100) ) the discontinuous lazy KRR predictor \eqref{SK},  (iv -HAN-SK(100)) the continuous lazy 
KRR predictor \eqref{DeCadix} in appendix. The linear head is the standard baseline used in transfer learning, while the MLP adds parametric capacity to adapt 
to the target task; both require training. By contrast, the lazy KRR readout \eqref{SK} is nonparametric and exploits the geometry of the learned feature space 
without any parameter training, whereas the continuous KRR-readout requires a small training overhead.

\begin{figure}[t]
    \centering
    \begin{subfigure}[t]{0.48\linewidth}
        \centering
        \includegraphics[width=\linewidth]{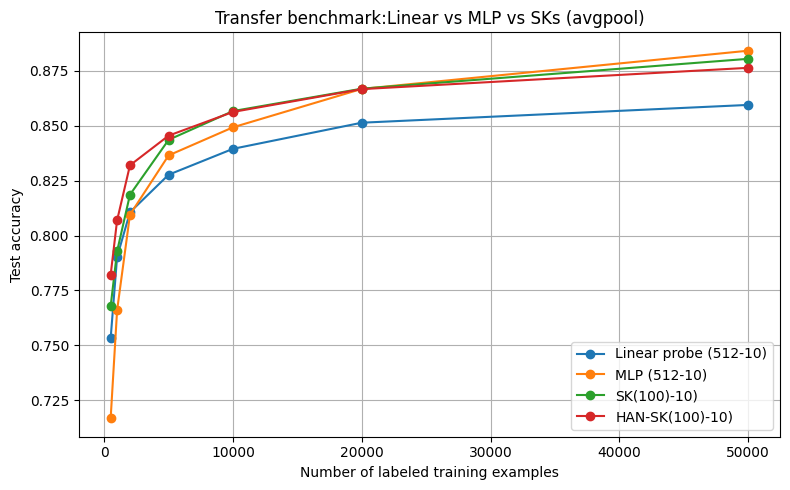}
        \caption{Test accuracy.}
    \end{subfigure}
    \hfill
    \begin{subfigure}[t]{0.48\linewidth}
        \centering
        \includegraphics[width=\linewidth]{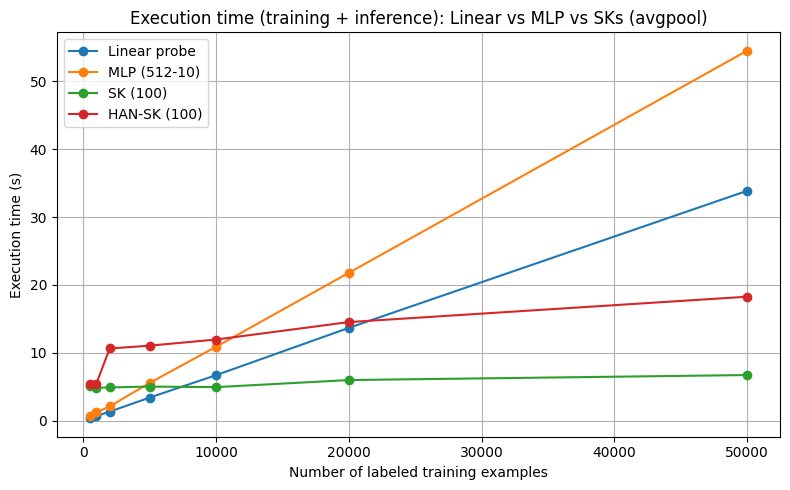}
        \caption{Wall-clock execution time.}
    \end{subfigure}
    \caption{Transfer learning on CIFAR-10 using a frozen, ImageNet-pretrained ResNet-18 backbone. Comparison of four readouts---linear head, MLP, discontinuous 
and continuous lazy KRR---across budgets of labeled CIFAR-10 examples.}
    \label{fig:transfer}
\end{figure}

Figure~\ref{fig:transfer} reports the resulting classification accuracy. The linear head provides competitive performance, confirming the quality of the 
pretrained ResNet-18 features. The MLP delivers moderate improvements, suggesting that limited nonlinear adaptation is beneficial. The lazy KRR readout reaches 
strong performance without any parametric training, indicating that it can effectively exploit the structure of the pretrained representations. These findings 
support the view that nonparametric readouts offer a simple yet powerful alternative for transfer learning. The continuous sparse readout (HAN-SK) is 
competitive in the small-to-medium budget regime but offers no clear advantage at full scale.

\subsection{Probing}\label{Probing}
We investigate the effect of probing~\cite{Bengio:2016,Hewitt:2019} at different depths for two representative architectures on CIFAR-10: a VGG-19 convolutional 
network and a vision transformer (ViT)~\cite{Dosovitskiy:2020}, each decomposed into five sequential layers. For both models, we construct four probes by 
progressively discarding the top layers and applying a readout on the resulting intermediate representations. This protocol allows us to quantify how predictive 
information evolves across depth.

\begin{figure}[t]
    \centering
    \begin{subfigure}[t]{0.48\linewidth}
        \centering
        \includegraphics[width=0.8\linewidth]{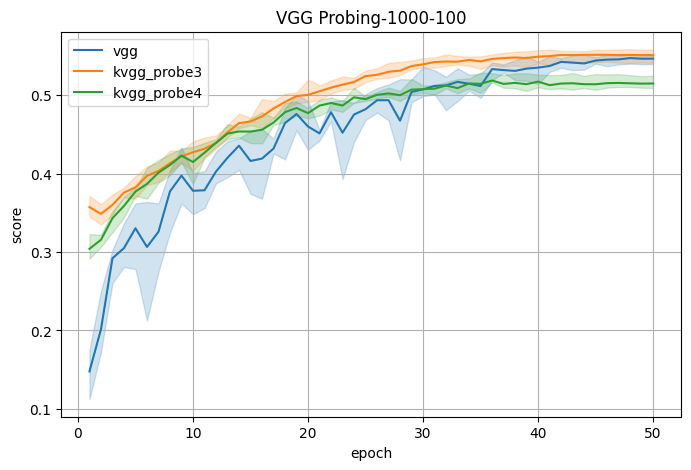}
        \caption{VGG-19 probing results (dataset 1000)}
    \end{subfigure}
    \hfill
    \begin{subfigure}[t]{0.48\linewidth}
        \centering
        \includegraphics[width=0.8\linewidth]{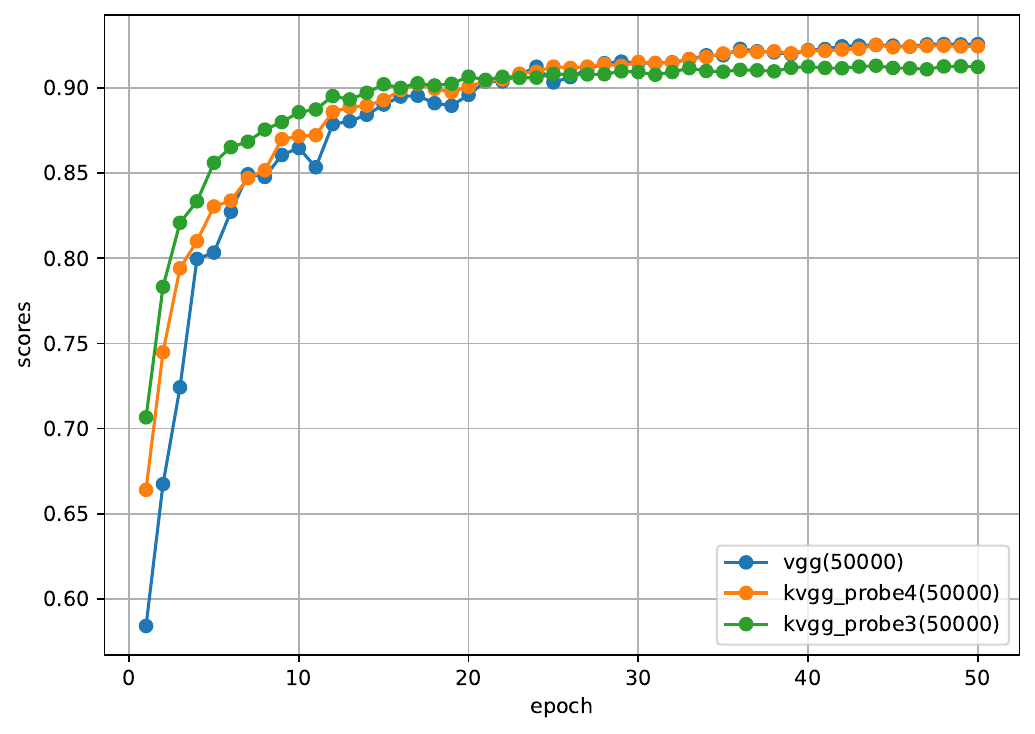}
        \caption{VGG-19 probing results (dataset 50000)}
    \end{subfigure}
    \vspace{1em}    
    \centering
    \begin{subfigure}[t]{0.48\linewidth}
        \centering
        \includegraphics[width=0.8\linewidth]{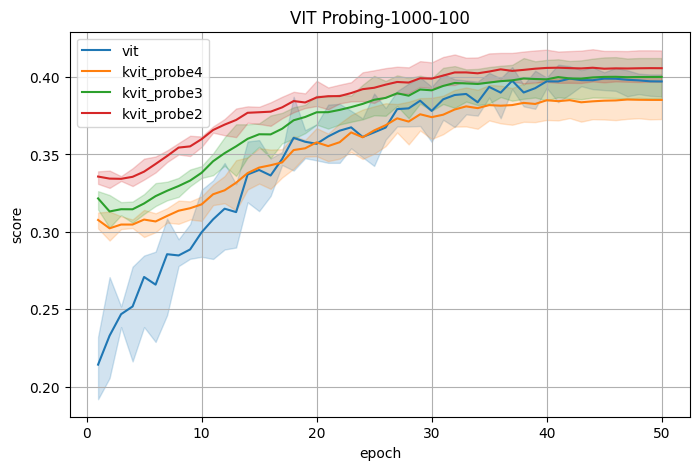}
        \caption{ViT probing results (dataset 1000)}
    \end{subfigure}
    \hfill
    \begin{subfigure}[t]{0.48\linewidth}
        \centering
        \includegraphics[width=0.8\linewidth]{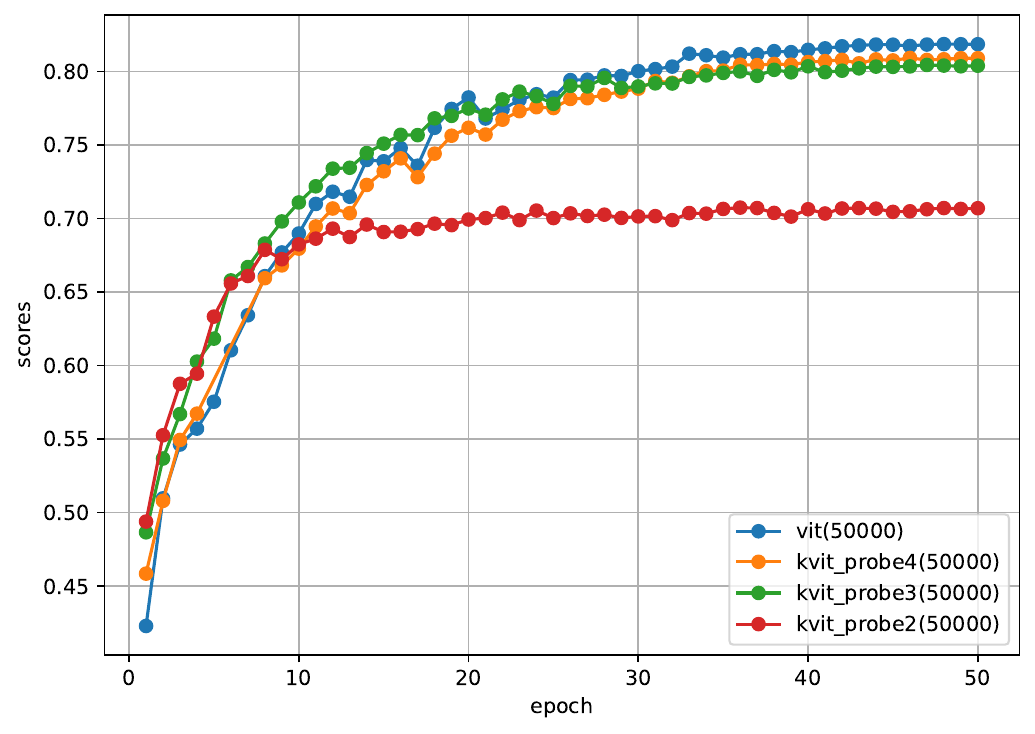}
        \caption{ViT probing results (dataset 50000)}
    \end{subfigure}
   
    \caption{Test accuracy (in \%) obtained by probing at different depths for VGG-19 and Vision Transformer on CIFAR-10. Each probe removes the top layers and 
evaluates the predictive power of intermediate representations. Top row: VGG-19; bottom row: ViT. Left column: 1{,}000-sample training subset (five run, average 
and min-max); right column: full 50{,}000-sample training set (one run).}
    \label{fig:probing}
\end{figure}

Figure~\ref{fig:probing} reports the test accuracy obtained for each probing configuration. Across both architectures, we observe that intermediate layers tend 
to yield stronger performance than the final layer. In particular, probing the last layer leads to a drop in accuracy for ViT, whose head consists of a linear 
layer, compared to the best intermediate layer. For VGG-19, this loss is less pronounced, the head being a multi-layer perceptron. The peak performance is 
typically achieved when removing one or two top layers, suggesting that these layers may over-specialize to the original training objective. While the overall 
trends are consistent, the magnitude of the improvement varies across models and probes, indicating that the benefit of probing depends on both the architecture 
and the readout.
These results suggest that intermediate representations provide a robust basis for probing and transfer with kernels, while final layers may discard information 
useful for alternative prediction tasks.

\subsection{Learning with probes}\label{Learning}
In this experiment we use the kernel-differentiable machinery of Section~\ref{Reminder-on-RKHS-methods} to \emph{train} the probes introduced in 
Section~\ref{Probing}. We use the AdamW optimizer~\cite{loshchilov2019adamw} with a cross-entropy loss.

\begin{figure}[t]
    \centering
    \begin{subfigure}[t]{0.48\linewidth}
        \centering
        \includegraphics[width=0.8\linewidth]{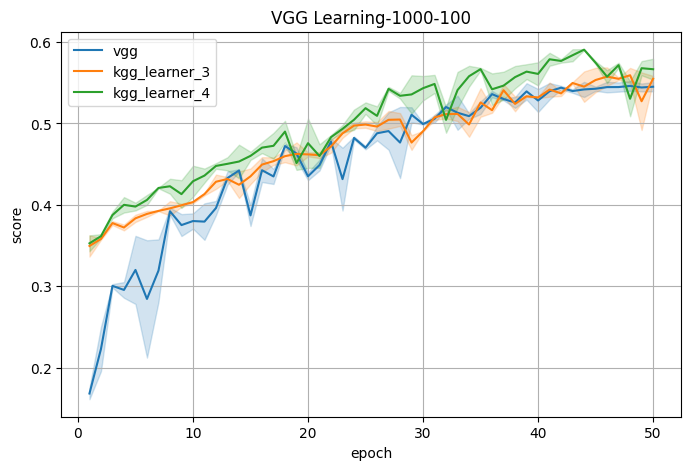}
        \caption{VGG-19 learning results (dataset 1000)}
    \end{subfigure}
    \hfill
    \begin{subfigure}[t]{0.48\linewidth}
        \centering
        \includegraphics[width=0.8\linewidth]{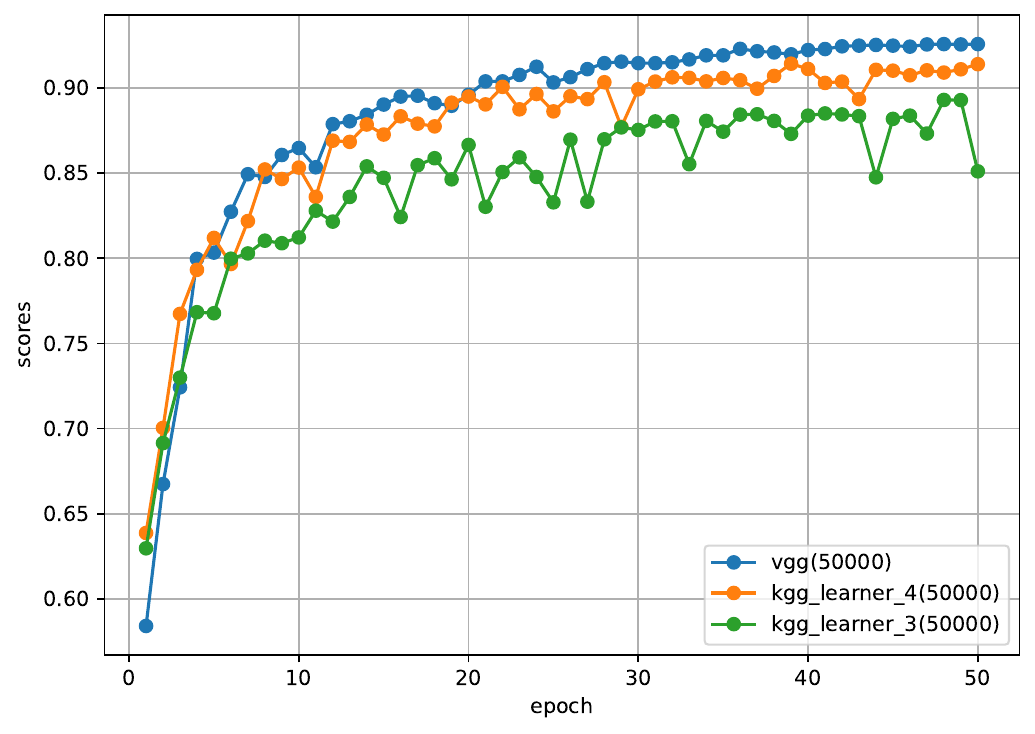}
        \caption{VGG-19 learning results (dataset 50000)}
    \end{subfigure}
    \vspace{1em}    
    \centering
    \begin{subfigure}[t]{0.48\linewidth}
        \centering
        \includegraphics[width=0.8\linewidth]{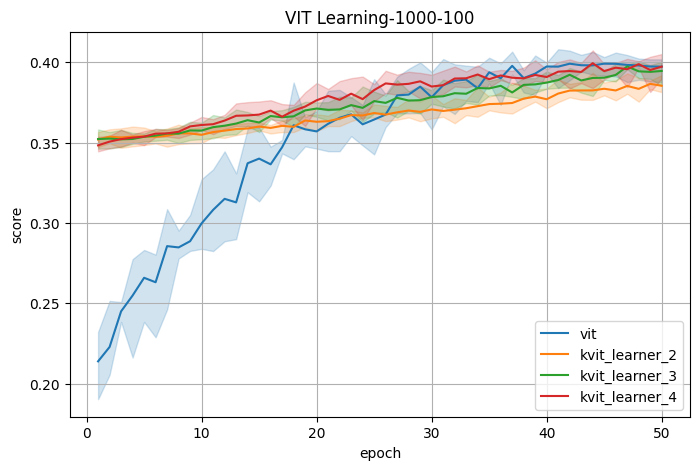}
        \caption{ViT learning results (dataset 1000)}
    \end{subfigure}
    \hfill
    \begin{subfigure}[t]{0.48\linewidth}
        \centering
        \includegraphics[width=0.8\linewidth]{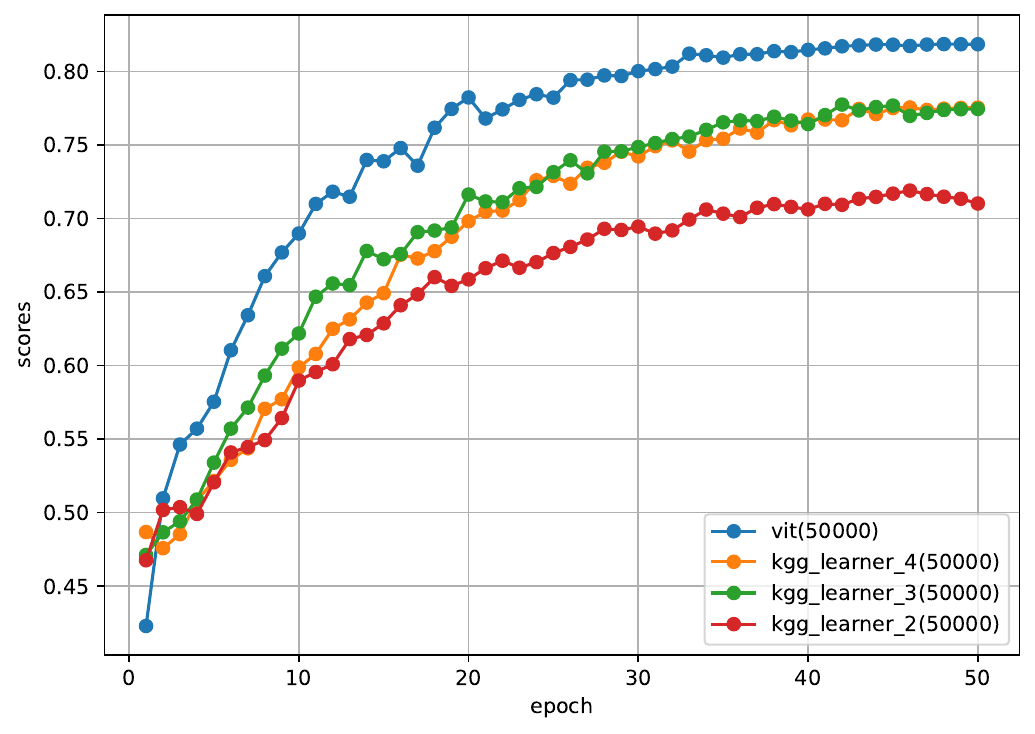}
        \caption{ViT learning results (dataset 50000)}
    \end{subfigure}
   
    \caption{Test accuracy (\%) when training kernel-based predictors at different depths for VGG-19 and ViT on CIFAR-10. Each configuration removes the topmost 
blocks and trains a kernel readout on the resulting representation. Top row: VGG-19; bottom row: ViT. Left column: 1{,}000-sample training subset (five run, 
average and min-max); right column: full 50{,}000-sample training set (one run).}
    \label{fig:learning}
\end{figure}

Figure~\ref{fig:learning} reports the resulting test accuracy on CIFAR-10. The trends closely mirror those observed in the probing experiments. For both 
architectures, training on intermediate representations often yields higher accuracy than training on the final layer; the best results are typically obtained 
after removing one or two top blocks, indicating that they are overly specialized to the original parametric classifier and are not well suited for kernel-based 
readouts.
Together with the probing results, these findings suggest that kernel methods are sensitive to the choice of representation and might benefit from features that 
are not overspecialized to a particular parametric head: effective kernel-based learning may require selecting or adapting intermediate representations rather 
than directly reusing the final layer of pretrained models. The same backend tools and training settings are used across both benchmarked architectures.

\subsection{Reinforcement learning}\label{ReinforcementLearning}

This experiment illustrates how the kernel-differentiable machinery can augment an existing reinforcement learning algorithm---namely, Double 
DQN~\cite{DQN:2016}---on the LunarLander environment. Double DQN is an online, off-policy, temporal-difference algorithm that estimates the action-value 
function. Let $(s_t, a_t, r_t, s_{t+1}) \in \mathcal{D}$ be transitions sampled from a replay buffer $\mathcal{D}$, where $s_t \in \RR^{S}$ is the state, $a_t 
\in \mathcal{A}$ the discrete action, with $|\mathcal{A}| = A$ the number of possible choices, and $r_t \in \RR$ the reward. The training objective is
\begin{equation} \label{eq:DQN}
 \inf_\theta \; \mathbb{E}_{\mathcal{D}} \Big[ \mathcal{L}\big(r_t +
 \gamma \max_{a'} Q(s_{t+1}, a'; \theta^-) - Q(s_t, a_t; \theta)\big) \Big],
\end{equation}
where $\gamma$ is the discount factor, $\theta$ the parameters of the online network, $\theta^-$ those of the target network, and $\mathcal{L}$ a loss function. 
The original DQN model uses
\begin{equation}
 Q(s, a; \theta) = \mathrm{ReLU}\big(\mathrm{ReLU}(s\,\theta_1)\,\theta_2\big)\, \theta_3, \qquad
 \theta_1 \in \RR^{S \times L},\
 \theta_2 \in \RR^{L \times L},\
 \theta_3 \in \RR^{L \times A},
\end{equation}
where $L$ is the latent dimension.
We refine this model by adding a kernel perturbation to the first and final layer:
\begin{equation}
 Q_k(s, a; \theta) = \mathrm{ReLU}\big(\big[\mathrm{ReLU}(s\,\theta_1) + \mathcal{P}_{k_1}(x_1,y_1)(s)\big]\,\theta_2\big)\, \theta_3 + 
\mathcal{P}_{k_3}(x_3,y_3)(s),
\end{equation}
where $\mathcal{P}_k(x,y)(\cdot)$ is the kernel ridge regressor introduced in Section~\ref{Dense-Kernel-Ridge-Regression}, adding the parameter sets $x_1 \in 
\RR^{B \times S}$, $y_1\in \RR^{B \times L}$, $x_3 \in \RR^{B \times L}$, $y_3 \in \RR^{B}$, where $B$ is the number of learnable kernel centers per 
module---i.e., the size of the support set $(x_i,y_i)$ on which the kernel ridge regressor $\mathcal{P}_{k_i}(x_i,y_i)$ is built. The augmented model is trained 
with the same algorithm and the same loss~\eqref{eq:DQN}, treating the additional kernel parameters $x_i, y_i$, $i=1,3$, as learnable parameters.

\begin{figure}[t]
    \centering
    \begin{subfigure}[t]{0.48\linewidth}
        \centering
        \includegraphics[width=\linewidth]{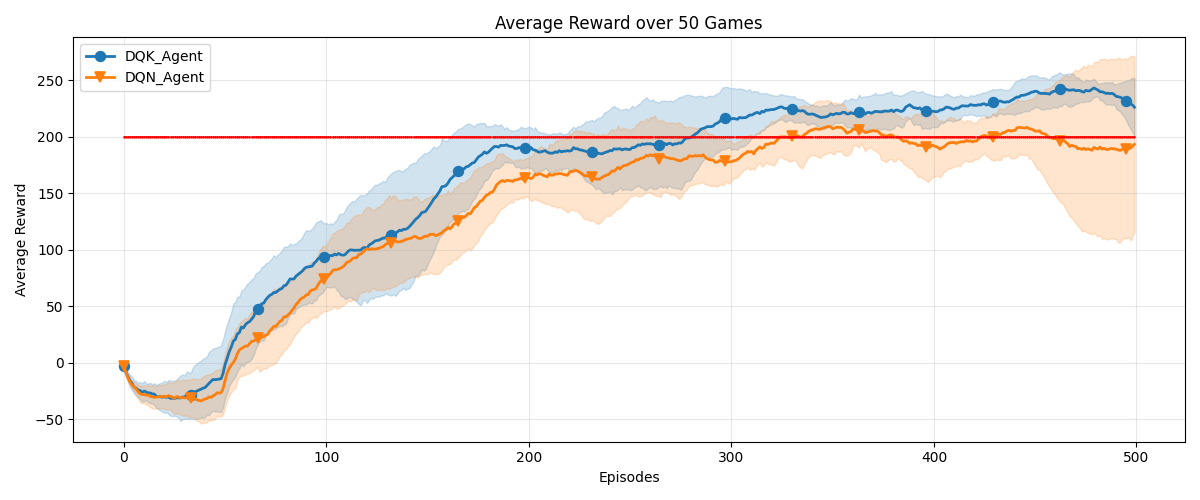}
        \caption{Episode rewards.}
    \end{subfigure}
    \hfill
    \begin{subfigure}[t]{0.48\linewidth}
        \centering
        \includegraphics[width=\linewidth]{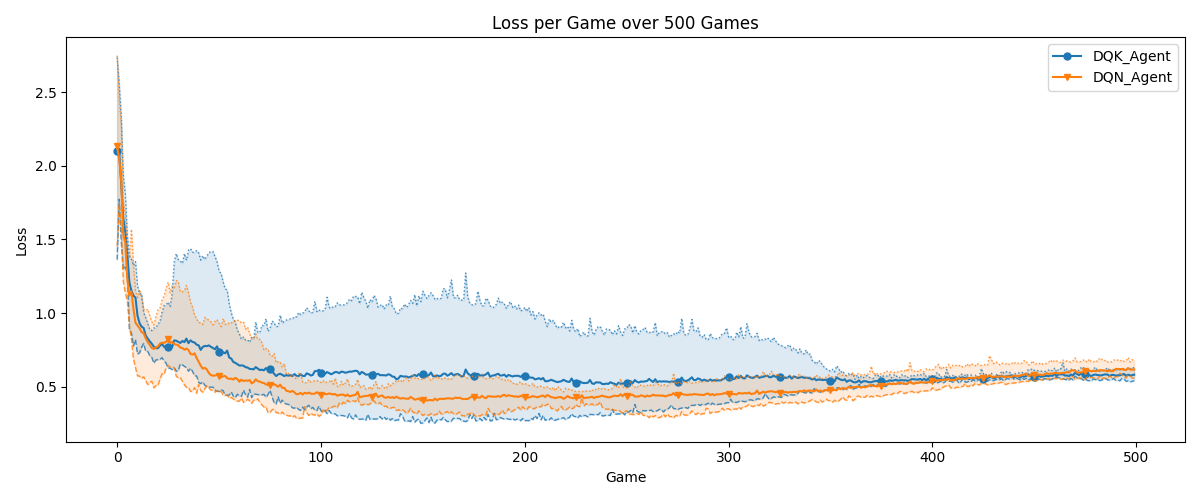}
        \caption{Training losses.}
    \end{subfigure}
    \caption{Rewards and losses for the original Double DQN (DQN\_Agent) and the kernel-augmented variant (DQK\_Agent) on the LunarLander Gymnasium environment. 
Curves show the mean and min--max range over five independent runs of $500$ episodes averaged over $50$ games.}
    \label{fig:LunarLander}
\end{figure}

Figure~\ref{fig:LunarLander} reports the reward and loss curves, averaged over 50 games and across five independent runs of $500$ episodes, with shaded min--max 
ranges. The kernel-augmented agent (DQK\_Agent) reaches higher rewards faster than the baseline (DQN\_Agent). The kernel-augmented agent exhibits higher TD 
losses during exploration; we interpret this as the additional kernel capacity fitting the TD targets less tightly during exploration while yielding a 
better-calibrated greedy policy.

\section{Conclusion}\label{Conclusion}

Our experiments support three main observations. First, on top of frozen pretrained features, lazy KRR readouts match or exceed parametric heads (linear and 
MLP) without any parametric training (Figure~\ref{fig:transfer}). Second, intermediate representations are systematically more amenable to kernel readouts than 
the final layer, both for fixed (probing, Figure~\ref{fig:probing}) and trained (Figure~\ref{fig:learning}) configurations, suggesting that final layers overfit 
to their original parametric head. Third, even in a setting as different from supervised learning as Double DQN, treating the targets and evaluation points of a 
kernel module as learnable parameters yields a measurable improvement at minimal implementation cost (Figure~\ref{fig:LunarLander}).

\paragraph{Limitations.} 

\textbf{Scale of evaluation.} Our experiments are run on a single CPU machine and use CIFAR-10 and LunarLander. We have not yet evaluated SKs on ImageNet-scale 
benchmarks, large language models, or diffusion models. The cubic-in-$M$ cost per query is mild for small $M$, but the practical wall-clock behavior on GPU---in 
particular, batched solves of $M\times M$ systems and $M$-NN lookup---remains to be benchmarked.

\textbf{Kernel and bandwidth selection.} The neighborhood size $M$ and the kernel function $\varphi$ act as hyperparameters. We did not perform an extensive 
sensitivity study; principled, data-driven choices of $M$ and $\varphi$ are an interesting direction.  The default sparse construction is globally 
discontinuous. Although it performs well empirically, applications requiring smooth predictors (e.g., gradient-based downstream optimization) may benefit from 
the continuous variants outlined in Appendix~\ref{HierarchicalContinuousSparseKernelRegressors}.

\textbf{Learnable evaluation points.} Treating evaluation points as learnable parameters may require additional regularization (e.g., the 
optimal-transport-based penalty discussed in Appendix~\ref{OT}); a thorough study is left to future work.

\paragraph{Summary.}
We introduced \emph{sparse kernels}, a differentiable, lazy variant of localized kernel ridge regression, and integrated them as modular components of standard 
deep learning pipelines. The construction makes explicit a three-fold parameterization of kernel methods---features, targets, and evaluation points---that can 
be selectively fixed or learned. Across transfer learning, probing, and a reinforcement learning case study, sparse kernels match or exceed neural counterparts 
while reducing training requirements.



\appendix
\section{Appendix}

This appendix collects experimental details (Section~\ref{NumericalExperimentSettings}) and additional methodological constructions 
(Sections~\ref{GreedySelection}--\ref{HierarchicalContinuousSparseKernelRegressors}) that complement the main text.

\subsection{Experimental details}\label{NumericalExperimentSettings}

All experiments are run on a CPU-only laptop\footnote{AMD 7950X3D processor 
with 64\,GB of RAM.}. Across all experiments, we use the kernel
$k(x,y) = \exp(-|S(x) - S(y)|_2)$, where $S(\cdot)$ is a standard normalizing 
map provided by the kernel library of~\cite{LeMeMi:2025}. We work in the interpolation regime, in practice setting the Tikhonov regularization parameter to 
$10^{-9}$. The sparse kernel 
implementation relies on FAISS~\footnote{\url{https://ai.meta.com/tools/faiss/}} 
for $M$-NN lookup.

\paragraph{Transfer learning (Section~\ref{Transfer}).}
We use a ResNet-18 pretrained on ImageNet\footnote{Weights available at 
\url{https://download.pytorch.org/models/resnet18-f37072fd.pth}.} as a frozen 
feature extractor, and transfer to the CIFAR-10 dataset\footnote{Loaded via 
\texttt{torchvision.datasets.CIFAR10}: 
\url{https://docs.pytorch.org/vision/main/generated/torchvision.datasets.CIFAR10.html}.}. 
The linear head is a single $512 \to 10$ linear layer; the MLP head is 
$512 \to 512 \to \text{ReLU} \to 10$. Both are trained with AdamW using a 
cross-entropy loss and a learning rate of $10^{-3}$. The discontinuous sparse kernel readout 
uses a bandwidth $M = 100$. The continuous sparse kernel readout 
uses a bandwidth $M = 100$, and a coarse dense kernel of size $\text{Dim}(H_{k,\sigma^0}) = 1000$, see \eqref{DeCadix}.

\paragraph{Probing (Section~\ref{Probing}).}
We use the CIFAR-10 dataset. The Vision Transformer is a minor adaptation of 
the PyTorch ViT-B/16 implementation\footnote{\url{https://github.com/pytorch/vision/blob/main/torchvision/models/vision_transformer.py}.} 
to the two-layer framework~\eqref{NN}; the VGG-19 convolutional network is 
adapted similarly\footnote{\url{https://docs.pytorch.org/vision/main/models/generated/torchvision.models.vgg19.html}.}. 
The sparse kernel readout uses bandwidth $M = 100$. Only one run is reported per curve on the full dataset for computational reasons.

\paragraph{Learning with kernel probes (Section~\ref{Learning}).}
The setting matches the probing experiment, with the kernel readout trained 
end-to-end. We use AdamW with batch size $64$, learning rate $10^{-3}$, no 
gradient scaler, and a cosine annealing schedule. Only one run is reported per curve on the full dataset for computational reasons.

\paragraph{Reinforcement learning (Section~\ref{ReinforcementLearning}).}
We use the LunarLander-v3 environment from the Gymnasium 
library\footnote{Maintained by the Farama Foundation: 
\url{https://gymnasium.farama.org/}.}. Hyperparameters: discount factor 
$\gamma = 0.99$, smooth $L^1$ loss $\mathcal{L}$, hidden dimension $L = 64$, 
state dimension $S = 8$, action space size $A = 4$, number of kernel centers $B = 64$. 
AdamW is run with batch size $64$, learning rate $10^{-3}$, no gradient scaler, 
and no scheduler. Results are averaged over five runs.

\begin{table}[h]
\centering
\caption{Hyperparameters across experiments.}
\begin{tabular}{lcccc}
\toprule
 & Transfer & Probing & Learning & RL \\
\midrule
Optimizer    & AdamW & ---   & AdamW & AdamW \\
Learning rate & $10^{-3}$ & --- & $10^{-3}$ & $10^{-3}$ \\
Batch size   & 128     & ---   & 64    & 64 \\
Bandwidth $M$ & 100  & 100   & 100   & --- \\
Scheduler    & ---   & ---   & cosine & --- \\
Epochs    & ---   & 50   & 50 & one by game move \\
\bottomrule
\end{tabular}
\end{table}

\subsection{Sparse kernel error estimates}\label{SparseKernelErrorEstimates}

This section provides a simple example of adapting RKHS error estimates to a localized sparse kernel regression, illustrating the basic mechanism by which other 
error estimates may be transferred to the sparse setting. To this end, we consider as an example the standard pointwise bound for the kernel ridge regressor in 
the interpolation regime, given by the power function. This error estimate is expressed in terms of basis functions; let $x=(x^1,\dots,x^{N})$, let 
$f(\cdot)\in\mathcal H_k$ be a continuous function, and denote by $f_k(\cdot)=k(\cdot,x)k(x,x)^{-1}f(x)$ the kernel ridge regressor of $f$. Then, considering 
normalized kernels (i.e.\ $k(z,z)=1$),
\be \label{EB}
   | f_k(\cdot) - f(\cdot) | \le \epsilon_k\big(\cdot\big)\, \|y(\cdot)\|_{\mathcal{H}_{k}}, \qquad \epsilon_k\big(\cdot\big) = \sqrt{ 1 - k(\cdot,x)  
\psi_k(x,\cdot) },
\ee
where $\psi_k(\cdot,x)$ are the basis functions~\eqref{psi}.
This estimate is numerically tractable and follows from~\cite{Caponnetto:2007} together with the reproducing property~\cite{Scholkopf:2002}. 
On each cell, the following \emph{local} error bound holds as a consequence of~\eqref{EB}:
\be \label{EBT}
   | f_k(\cdot) - f(\cdot) | \le \epsilon_k^\sigma\big(\cdot\big)\, \|y(\cdot)\|_{\mathcal{H}_{k}},\qquad  \epsilon_k^\sigma(\cdot) = \sqrt{ 1 - 
k(\cdot,x^\sigma)  \psi_k(x^\sigma,\cdot) }.
\ee

\subsection{Greedy selection}\label{GreedySelection}
In some settings one wishes to reduce the computational cost of dense or sparse KRR. A simple approach is to select a representative subsequence of $N' < N$ 
points from the dataset $(x^1,\dots,x^N)$ by means of an incremental nested sequence $X^{n} \subset X^{n+1} \subset \dots \subset X^N$, and the following greedy 
rule\footnote{Often referred to as the Gonzalez algorithm; see, e.g., 
\url{https://www.cs.umd.edu/class/spring2025/cmsc451-0101/Lects/lect06-greedy-k-center.pdf}.}:
\be
\label{GS=2}
X^{n+1} = X^n \cup \arg\sup_{x' \in \{x^{n+1},\dots,x^N\}} d(x',X^n),
\ee
where $d(x',X^n) = \inf_{x \in X^n} d(x',x)$ and $d$ is a user-defined distance.

\subsection{Optimal transport error loss function}\label{OT}

Optimal-transport losses produce continuous \emph{bijective maps} that can match discrete features $x \mapsto y$ up to a permutation $\sigma \in \Sigma$ with 
kernel machines. Suppose first that $(x,y)$ live in the same space $\RR^D$. For a local cost $c(\cdot,\cdot)$, consider the transport functional
\be
\label{LSAP-deux}
M(x,y) = \sum_{n=1}^N c\big(x^n,y^{n}\big).
\ee
Minimizing it over permutations, $\overline{\sigma} = \arg\inf_{\sigma \in \Sigma} \sum_n c(x^n,y^{\sigma^n})$, recovers the $c$-convex \emph{Monge} problem, 
which can be solved by linear-sum-assignment algorithms.

The \emph{Gromov--Monge} functional generalizes this to unrelated spaces $D_x \neq D_y$ via two cost functions $c_x,c_y$ (e.g., Euclidean):
\be
\label{GM-discrete}
GM(x,y) = \sum_{i,j=1}^N \big|c_x(x^i, x^j) - c_y(y^i, y^j)\big|^2.
\ee
This functional is a natural candidate to add to a base loss, $\mathcal{L}(\theta) + GM(\theta_x,\theta_y)$, in order to encourage a balanced spatial 
distribution of points whenever the features $\theta_x$, the targets $\theta_y$, or both, are learnable parameters of a kernel ridge regressor as in~\eqref{2L}.

\subsection{Hierarchical, continuous sparse kernel regressors} \label{HierarchicalContinuousSparseKernelRegressors}

As noted in the main text, the space $\mathcal{H}_{k}^M$ contains functions that are continuous on each cell $\Omega^\sigma$ but possibly discontinuous 
globally, because of the discontinuity of $\mathbf{1}_{\Omega^\sigma}(z)$. When global continuity is required, we describe two alternatives; the first 
follows~\cite{Han:2022}.

To obtain a \emph{linear} globally continuous model class, one introduces a continuous partition of unity, namely a family of continuous functions 
$(w^\sigma)_{\sigma \in \Sigma}$ subordinate to the tessellation (locally finite overlap, $\sum_\sigma w^\sigma(z)=1$), and considers
\be
\mathcal{H}_{k}^M = \Bigl\{
y_k(\cdot) = \sum_\sigma w^\sigma(\cdot)\, y^\sigma(\cdot),\quad y^\sigma \in H_{k,x^\sigma}\Bigr\},
\ee
equipped with the induced (image/quotient) norm
\[
\|y\|_{\mathcal{H}_{k}^M}^2 \coloneqq \inf\Bigl\{\sum_\sigma \|y^\sigma\|_{\mathcal{H}_{k,x^\sigma}}^2 : y=\sum_\sigma w^\sigma y^\sigma\Bigr\}.
\]
Under mild boundedness and local finiteness conditions, $\mathcal{H}_{k}^M$ is a Hilbert space.
The resulting blended predictor $y_k(z)=\sum_\sigma w^\sigma(z) y^\sigma(z)$ is globally defined and avoids discontinuities at cell boundaries. A typical choice 
for the weights is the sparse Nadaraya--Watson form
$$
w^\sigma(\cdot) = \frac{\varphi(d(\cdot, \Omega^\sigma))}{\sum_{\tau \in \Sigma} \varphi(d(\cdot, \Omega^\tau))},
$$
for some positive function $\varphi$, where $d(\cdot,\Omega^\sigma)$ is a distance-type function, e.g., $d(z,\Omega^\sigma) = \inf_n |z - x^{\sigma^n}|$ or 
simply $|z - x^{\sigma^1}|$ as in~\cite{Han:2022}.

A second way to obtain a globally continuous regressor is to use a positive definite kernel $k$ that vanishes on cell boundaries, i.e., $k^\sigma(x,y) = 0$ 
whenever $x$ or $y$ lies on the boundary $\Gamma^\sigma$ of $\Omega^\sigma$. Denoting by $H_{k,\sigma}^0$ the resulting RKHS on $\Omega^\sigma$, we introduce 
the following hierarchical sparse construction:
\be \label{DeCadix}
\mathcal{H}_{k}^M = \Bigl\{
y_k(\cdot) = y^0(\cdot) + \sum_{\sigma \in \Sigma} \mathbf{1}_{\Omega^\sigma}(\cdot)\, y^\sigma(\cdot),\quad y^\sigma \in H_{k,\sigma}^0,\ y^0 \in 
H_{k,\sigma^0}\Bigr\},
\ee
where $H_{k,\sigma^0}$ is a global, dense, coarse RKHS as in Section~\ref{Dense-Kernel-Ridge-Regression}, built from a small subset $x^{\sigma^0}$ of selected 
points---obtained, for instance, via the greedy procedure of~\eqref{GS=2}.

\end{document}